\documentclass[12pt]{article}

\usepackage{algpseudocode}
\usepackage{booktabs}
\usepackage{array}
\usepackage{tabularx}
\usepackage[utf8]{inputenc}
\usepackage{amssymb,amsmath}
\usepackage{amsfonts}
\usepackage{latexsym}
\usepackage{geometry}
\usepackage[francais,english]{babel}
\usepackage[T1]{fontenc}
\usepackage{eepic}
\usepackage{epic}
\usepackage{epsfig}
\usepackage{epsfig}
\usepackage{graphics,color}
\usepackage{verbatim}
\usepackage{lscape}
\usepackage{amsmath}
\usepackage{amssymb}
\usepackage{amsthm}
\usepackage{amsmath}
\usepackage{placeins}
\usepackage{float}
\usepackage{caption}
\usepackage{color}
\usepackage{setspace}
\usepackage{frcursive}
\usepackage{amsfonts}
\usepackage[linesnumbered, ruled]{algorithm2e}
\usepackage{url}
\usepackage{hyperref}
\usepackage[dvipsnames]{xcolor}

\makeatletter
\renewcommand\paragraph{\@startsection{paragraph}{3}{\z@}%
            {-2.5ex\@plus -1ex \@minus -.25ex}%
            {1.25ex \@plus .25ex}%
            {\normalfont\normalsize\bfseries}}
\makeatother
\usepackage{mathrsfs}
\usepackage{soul}
\usepackage{tikz}
\usepackage{tikz,fullpage}
\usetikzlibrary{arrows,petri,topaths,automata}
\usepackage{tkz-berge}
\usepackage{pgfplots}

\makeatletter

\usepackage{tikz}
\usepackage{tkz-berge}
\usepackage{tikz,fullpage}
\usepackage{url}

\newif\ifbold
\boldfalse
\boldtrue

\newcommand{\bbf}{\ifbold\bgroup\bf\fi}
\newcommand{\ebf}{\ifbold\egroup\fi}

\renewcommand{\textbf}[1]{\begingroup\bfseries\mathversion{bold}#1\endgroup}

\makeatletter    
\renewcommand{\section}{\@startsection {section}{1}{\z@}%
             {-2ex \@plus -1ex \@minus -.2ex}%
             {1ex \@plus.2ex}%
             {\normalfont\Large\rmfamily\bfseries}}

\renewcommand{\subsection}{\@startsection{subsection}{2}{\z@}%
             {-1.25ex\@plus -1ex \@minus -.2ex}%
             {.75ex \@plus .2ex}%
             {\normalfont\large\rmfamily\bfseries}}

\def\@listI{\leftmargin\leftmargini       
            \parsep .25ex \@plus .1ex     
            \topsep .25ex \@plus .1ex     
            \itemsep \parsep}
\let\@listi\@listI
\@listi

\makeatother

\makeatletter

\@addtoreset{equation}{section}
\makeatother

\makeatletter

\@addtoreset{table}{section}
\makeatother

\definecolor{purple}{rgb}{0.4,0.2,1}

\title{
{\LARGE Research Paper: \\ \vspace{0.5cm}
\LARGE\bf A Column Generation based Heuristic for the Tail Assignment Problem }\\ \vspace{3cm}
}

\vspace{0.5cm} 

\author{
{
\large\bf SAMBREKAR AKASH$^1$, \large\bf ER RAQABI El Mehdi$^1$}\\
\\
1. Department of Mathematical and Industrial Engineering,\\
Polytechnique Montreal, H3T 1J4, Canada
}

\date{\bf \vfill \today}

\usepackage{graphicx}
\graphicspath{ {./Figures/} }

\begin{document}
\maketitle
\thispagestyle{empty}
\newpage
\setlength{\parindent}{4em}
\setlength{\parskip}{1em}
\renewcommand{\baselinestretch}{1.5}

\begin{abstract}
This article proposes an efficient heuristic in accelerating the column generation by parallel resolution of pricing problems for aircrafts in the tail assignment problem (TAP). The approach is able to achieve considerable improvement in resolution time for real life test instances from two major Indian air carriers. The different restrictions on individual aircraft for maintenance routing as per aviation regulatory bodies are considered in this paper. We also present a variable fixing heuristic to improve the integrality of the solution. The hybridization of constraint programming and column generation was substantial in accelerating the resolution process.\par
\vspace{0.4cm}
\noindent
\emph{Keywords}: Tail Assignment; Column generation; Constraint programming; Heuristic

\end{abstract}
\newpage

\section{INTRODUCTION}
\normalsize
\vspace{0.4cm}

\textbf{The airline business}. Since the availability of air transport, humans have been traveling intensively using planes. Such a mean of transport is allowing trips within the same continent and between continents. In parallel to this increase in demand, the airline industry have been evolving on many aspects. First, many companies emerged gradually in the industry. Some are targeting the international flights, i.e. between countries, while other are targeting the local flights, i.e. within the same country. Even on the local level, air transport seems to be competitive with other means of transport such as railway, cars, etc. With this growth in coverage, processes are continuously improved to reduce significantly costs, allow better flexibility for customers, and make the ticketing fully online. This ease in process is attracting more customers than before. Consequently, the increase in demand has led to, and will continue to lead, to large investments on many levels including the infrastructure, the fleet of planes, and the human resources. Furthermore, the increasing frequency of flights has made the planning and scheduling of planes usage more complex.  \par

\noindent
\textbf{The operational system}. Airlines companies operate on a point-to-point system, a hub-and-spoke system or a hybrid system. In the first one, illustrated in Fig. 1, flights go directly from the origin to the destination. In the second one, illustrated in Fig. 2, a big airport is selected to be the center of the hub and the flights stopover it before reaching the destination. The hub system may be composed of one hub or many hubs. In the hybrid system, there are both point-to-point flights and hub flights. The choice is based on many factors such as the operating costs as well as the demand frequency. Each plane must respect international regulations in order to fly over different places. In addition, it cannot stay more than 32 hours on the sky and must go through frequent maintenance processes once completed the cycle. After finishing a flight, it should at least remain 20 min on the ground before taking another flight. Based on the historical data, companies plan the utilization of the fleet on a monthly basis. \par

\begin{figure}[h]
  \centering
  \begin{minipage}[b]{0.4\textwidth}
    \includegraphics[width=\textwidth]{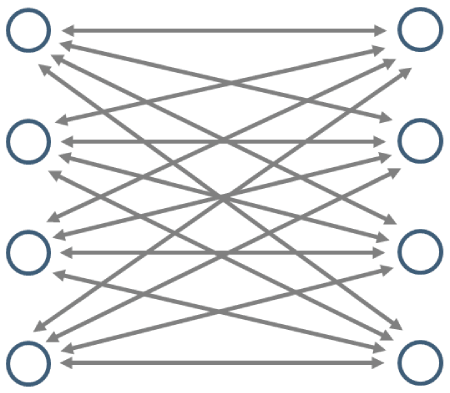}
    \caption{Point system}
  \end{minipage}
  \hfill
  \begin{minipage}[b]{0.4\textwidth}
    \includegraphics[width=\textwidth]{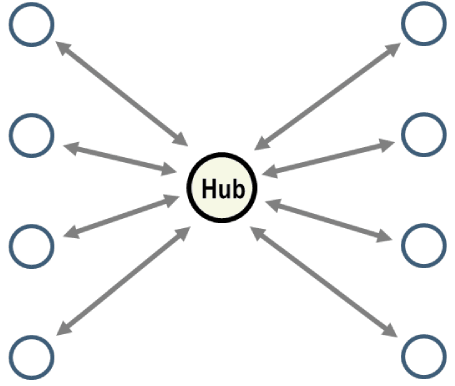}
    \caption{Hub system}
  \end{minipage}
\end{figure}

\noindent
\textbf{The planning complexity}. In terms of size and complexity, planning in the airline industry is considered one of the most difficult problems known. For a given time horizon, it has four main parts, see Fig. 3. The first part, i.e. flight scheduling, select a set of flights with fixed departure and arrival periods while seeking expected profit maximization. Then, fleet assignment is usually performed. It identifies the aircraft type to be assign to each scheduled flight based on available aircraft and capacity restrictions. Next, aircraft routing is tackled. Each aircraft is assigned to a set of flights while ensuring maintenance requirements. Finally, crew scheduling composed of crew pairing and crew rostering is solved. It assigns required crew personnel to each of the planned flights in the schedule and at the same time  satisfying  the  rules as per the airline regulatory bodies such as FAR, DGCA, etc. Recovery scenarios are also elaborated to anticipate unexpected events such as delays, accidents, sick members, and airport closures. In such a case, the aircraft routing problem must be resolved to handle these perturbations. Survey articles by Gopalan and Talluri \cite{gopalan1998mathematical} and Barnhart et al. \cite{barnhart2003applications} contain good overview of airline planning problems, and the use of operations research models to solve them. This research area has also a specific lexicon. A routing that starts and ends in the same airport is called a rotation. Similarly, a pairing may start and end in the same crew base, i.e. where they actually live, spans from one to five days and is sometimes called itinerary. The main two resources for such problems are the fleet size and the available crew. The main costs are fuel consumption and crew salaries. The goal is minimizing costs through an efficient usage of both airplanes and crew members while transporting as much customers as possible in each flight. The main constraints are international regulations and labour unions that must be respected in order to ensure long-term operations. \par

\begin{figure}[h]
\noindent
\includegraphics[width=\textwidth]{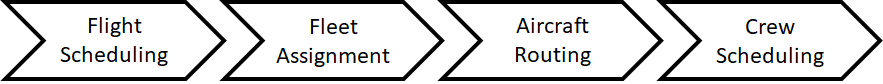}
\caption{Airline planning process}
\end{figure}

\noindent
\textbf{This research paper}. In this paper, we focus on solving the aircraft routing problem known in the literature \cite{gronkvist2005tail} as the \emph{Tail Assignment Problem} (TAP) for one of the largest airline companies in India. It incorporates all aspects of aircraft routing and maintenance requirements. When tackling this problem, the aim is to deal with the operational perturbations, maintain feasibility, and determine an optimal fleet schedule that minimizes costs and provides the sequences of flights assigned to each aircraft. This schedule must also satisfy maintenance constraints specific to each individual aircraft. 

\noindent
\textbf{The research purpose}. The purpose of this paper is twofold: (1) Present the tail assignment problem and a case study from an large airline company, and (2) Describe the mathematical formulation and the resolution approach that combines column generation (CG), constraint programming (CP), and heuristics. 
The research in this paper complements the available literature for this important problem by introducing a new approach to handle maintenance called maintenance assignment, a parallel resolution of the sub-problems obtained from Dantzig-Wolfe decomposition. The practical approach can act as a basis for other solution approaches that seek tackling large-scale instances of the problem.\par

\noindent
\textbf{The research organization}. The rest of the paper is organized as follows: A brief overview of the recent literature for the tail assignment problem is presented in Sect. 2. Section 3 is devoted to a detailed description of the problem, while the mathematical programming formulation is given in Sect. 4. The resolution approach as well as the heuristics implemented are described in Sect. 5. Section 6 presents the real-world cases tested and the associated computational results. Finally, concluding remarks follow in Sect. 7. \par

\section{LITERATURE REVIEW}
In this section, we will study the research conducted in the area of tail assignment and aircraft routing problems. We will also discuss the different optimization techniques implemented by researchers to solve these problems. Predominantly, the fleet assignment and aircraft routing problems have been studied extensively. But in recent years, TAP has garnered researchers' attention in both industry and academia given its ability to handle disruptions in the airline planning stages. The Aircraft Routing Problem (ARP) plans aircraft routes for specific aircraft types since aircrafts of same type follow the same routine of periodic maintenance checks. But ARP does not consider the individual maintenance regulations which vary with each aircraft. In \cite{desaulniers1997daily}, Desaulniers solves a daily aircraft routing problem for determining schedules for the aircraft's with provision of varying departure times of flights. They proposed a set partitioning model and time constrained multi-commodity network flow model. The measure of improving the robustness of the routes to disruptions is considered by several researchers. In \cite{liang2015robust}, Liang et al propose a weekly line of flight (LOF) network model with an objective to improve the  resilience to disruptions by providing adequate buffer times between connecting flights. Borndorfer \cite{borndorfer2010robust} constructs routes by considering disruptions from preceding days that could affect the following planned maintenance activity. In \cite{bacsdere2014operational}, Basdere and Bilge propose an integer linear programming model based on connection network to plan feasible routes for individual tails, which maximizes the utilization of the total remaining flying time of aircraft fleet. Sarac \cite{sarac2006branch} addresses the different maintenance checks mandated by the FAA in the aircraft routing problem. They considered the legal flying hours limits and restrictions at the maintenance bases. They resolve the problem using a Branch and Price framework with follow on rule for fixing the variables. Feo and Bard \cite{feo1989flight} propose a model to determine maintenance base locations and develop flight timetables that meet the demand for maintenance checks.\par

\noindent
Airlines prefer to follow the routes planned during ARP for better management but disruptions make it difficult to follow the plan. It is certain that LOF's have to be restructured to recover from disruptions and to satisfy operational constraints. These routes planned by ARP can be used as an input for TAP. Several researchers like Lapp and Maher have used the LOFs which are a priori generated by the ARP as an input for TAP. Lapp \cite{lapp2012incorporating} proposed a integer programming model to build LOFs such that maintenance reachability is maximized. Maher \cite{maher2018daily} proposed an iterative algorithm that quickly provides a feasible solution by reducing the maintenance mis-alignments for the given input of LOFs compared to the column generation approach but significant gap was evident from the optimal solution. Ruther \cite{ruther2013integrated} proposed an integrated approach for aircraft routing, crew pairing and tail assignment problem by resolving close to day of operations using column generation.

\noindent
In Literature, some researchers have integrated the ARP and TAP by constructing the routes for each individual aircraft. One way to solve such problems would be to enumerate all possible routes for each tail satisfying the individual aircraft constraints and later modeling it as a set-partitioning problem. Well, its not possible to generate all possible routes and solve such a large-scale problem with millions of variables. In such context, column generation can be employed for a dynamic generation of routes by resolving a resource constrained shortest path problem for each tail. Extensive work in the area of TAP was carried out by Gronviskt \cite{gronkvist2005tail} in his doctoral thesis. Gronviskt resolved TAP by combining constraint programming and column generation, which significantly reduced the resolution time due to reduced network size. He employed different techniques like Dual Re-evaluation scheme and suggested a randomized order of resolving the pricing problems to generate dissimilar columns in column generation iterations. Rather than implementing a complex branch and price heuristic, the paper suggests integer fixing heuristics to obtain an integer solution since the objective was to reach a good feasible solution as quickly as possible. Several researchers like Rousseau \cite{rousseau2004solving}, Gabteni \cite{gabteni2009combining} and Gualandi \cite{gualandi2009constraint} worked on hybridization of column generation and constraint programming to accelerate the convergence of resolution process.

\noindent
Since TAP is to be deployed at the operational level that handles immediate disruptions, the user expects resolution in few minutes. The paper presented here provides quicker resolution time to TAP by resolving the pricing problems in a parallel way while ensuring the selection of dissimilar columns. It also incorporates the LOFs planned during the ARP and restructures them if necessary to handle any disruptions. Based on our review of existing studies, we realized that only few studies related to  maintenance regulations like maintenance checks, assigning planned maintenance and creating maintenance opportunities along the route of the tail have been considered.

\section{PROBLEM DESCRIPTION}
The TAP includes determining the routes for a set of individual aircraft that are identified by tail numbers that should cover a set of flights from the planned schedule. The routes have to satisfy various actual operational constraints like flight connection constraints, maintenance constraints and activity restrictions. They are discussed in detail below. In our problem, we plan both flights and maintenance operations. Hence, we use the term activity to refer to both of them.

\subsection{Flight connection constraints}
If the arrival base of a first leg and departure base of a second leg is same and sufficient ground time is available, then it is possible to connect these two legs by a flight connection. Minimum ground time (MGT), also as called turn-around-time, is the minimum time required by an aircraft to be ready for its next take-off after it has just landed on a particular airport. The MGT varies considerably from 25 to 45 minutes depending upon the level of aircraft activity at a particular airport. Similarly, the airlines also have Maximum connection time (MCT) to restrict an aircraft from being idle for a long duration. 

\noindent
\subsection{Maintenance constraints} 
Maintenance constraints are mainly of two types: pre-assigned maintenance activity and periodic maintenance checks. The pre-assigned activities usually represent a maintenance activity specific to an individual tail to be carried out a particular maintenance base between specified times. This pre-assigned has to be necessarily assigned in the given time-frame with a flexibility in its start time. This activity is planned based on the recommendation of aircraft manufacturers, usually Airbus and Boeing, but sometimes, the maintenance and engineering team (M\&E) of airlines might be more restrictive in its maintenance planning. These maintenance activities may usually span for multiple days. On the other hand, periodic maintenance checks are not restricted to be done on a specific time-frame or on a particular base. In our problem, we consider two maintenance checks. The flying hours (FH) track the number of consecutive flight hours flown by an individual tail from the previous maintenance check. Similarly, flying Cycle (FC) tracks the consecutive number of landings made from the previous check. These maintenance checks vary depending on the type of aircraft. The Table below shows an example for the FH and FC checks for a specific tail. The tail number 485 needs a maintenance check before the tail has been used for consecutive 150 flying hours and 500 cycles. It cannot exceed these restrictions. Thus, these periodic maintenance checks have to be planned accordingly depending upon the maintenance opportunities available throughout the route of a specific tail. The periodic checks could be done at different maintenance bases depending upon the hangar facilities of the bases. \par

\begin{table}[ht]
\centering
\begingroup
\setlength{\tabcolsep}{10pt} 
\renewcommand{\arraystretch}{1.5} 
\begin{tabular}[t]{lcc}
\toprule
Tail number&FH(hours)&FC(cycles)\\
\midrule
Tail 485&150&500\\
Tail 979&350&250\\
\bottomrule
\end{tabular}
\endgroup
\caption{FH and FC checks example}
\label{table:1}
\end{table}%

\noindent
\subsection{Activity restrictions}
These restrictions forbid aircrafts from flying in certain sectors depending upon their qualification. For extended operation (ETOP) sectors, usually a twin engine is required. It may be that the specific aircraft does not have enough in-flight entertainment systems to serve certain sectors. Usually, some tails have runway restrictions on particular airport. There are also situations when certain routes/flight connections have to be assigned to a specific tail to adhere crew connections. \par

\noindent
\subsection{Retaining the rotation line} 
The TAP is an extension of aircraft routing problem but resolved at the operational level so as to handle any disruptions without changing the routes that were planned at the Aircraft Routing Problem (ARP). After the aircraft routing problem has been resolved and routes have been planned and assigned to an aircraft line, the TAP tries to assign the flights that were planned by the aircraft routing problem on the same aircraft line. To achieve this, we provide bonuses to the  generated routes that follow the same routes or sub-routes planned during the ARP.\par

\section{MODEL}
Referring to the description given in the previous section, the problem considered can be formulated using a mixed integer programming (MIP) model. The problem is subject to two sets of constraints: flight coverage constraints and tail constraints.

\subsection{Sets}
The sets used in the mathematical formulation are as follows:\\
\noindent
\textbf{${\cal{F}}$}: set of activities to be covered, $f \in {\cal{F}}$.\\
\noindent
\textbf{${\cal{G}}$}: set of routes generated, $g \in {\cal{G}}$. \\
\textbf{${\cal{T}}$}: set of tails/aircrafts, $t \in {\cal{T}}$. \\
{\bf V$_{t}$}: set routes valid for each tail, $t \in {\cal{T}}$.

\subsection{Parameters}
The TAP has parameters that are available a priori. This data is incorporated into the mathematical formulation as constants. They are written in bold style in the model.\\
\noindent
{\bf C$_{i}$}: penalty cost for uncovered activity $i \in {\cal{F}}$. \\
{\bf C$_{r}$}: cost of each route, $r \in {\cal{G}}$. \\
{\bf a$_{ir}$}: equals 1 if route $r \in {\cal{G}}$ covers activity $i \in {\cal{F}}$, 0 otherwise.

\subsection{Variables}
The decision variables considered in the TAP are as follows:

\noindent
{\bf f$_{i}$}: binary variable equal to $1$ if activity $i \in {\cal{F}}$ is uncovered, 0 otherwise. \\
{\bf x$_{r}$}: binary variable equal to $1$ if route $r \in {\cal{G}}$ is selected, 0 otherwise. \\

\subsection{Constraints}
The model constraints are as follows:

\noindent
\textbf{Flight coverage}: ensures that every activity has  to be covered only once or can be left uncovered.
\begin{equation}
\begin{split}
f_i + \sum_{r \in  {\cal{G}}} {\bf a_{ir}}.x_r = 1 \hspace{5mm} \forall i \in {\cal{F}}
\end{split}
\end{equation}

\noindent
\textbf{Tail constraint}: indicates that at most one route can be assigned to each tail.
\begin{equation}
\begin{split}
\sum_{r \in  {\cal{V}}_{t}} x_r \leq 1 \hspace{5mm} \forall t \in {\cal{T}}
\end{split}
\end{equation}

\noindent
\textbf{Binary restrictions}: on TAP variables {\bf f$_{i}$} and
{\bf x$_{r}$}.

\subsection{Objective function}
In the TAP, we seek minimizing the following objective function:
\begin{equation}
\begin{split}
{\bf Min} \hspace{2mm} \sum_{i \in  {\cal{F}}} {\bf C_{i}}.f_i + \sum_{r \in  {\cal{G}}} {\bf C_{r}}.x_r
\end{split}
\end{equation}

\section{RESOLUTION}
We solve TAP by column generation technique. In this technique, we decompose the problem into two components: a restricted master problem (RMP), which contains a restricted number of variables and a pricing problem (PP), which generates negative reduced cost routes for the tails. The resolution process is summarized in the Fig. 4. At first, we initialize the RMP and provide its dual solution to the pricing problems. The routes generated by pricing problems are added to the RMP and updated dual solution is found after resolving the RMP. The process is continued until there are no negative reduced cost routes. After the RMP converges, we start fixing few variables and connections to ensure that paths, which are better in a standpoint view of integrality, are generated. After fixing a significant part of the problem, we restore the integrality constraints on the model and solve it as a integer programming model.

\begin{figure}[h]
\centering
\noindent
\includegraphics[width=0.5\textwidth]{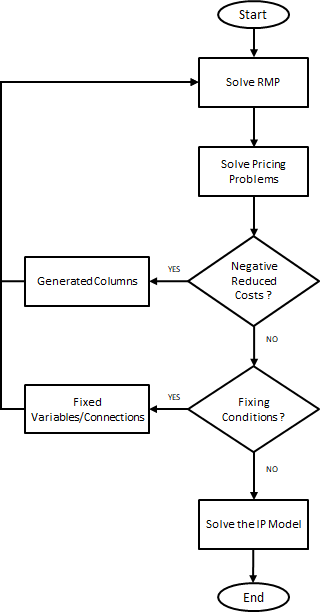}
\caption{Resolution approach}
\end{figure}

\subsection{Restricted Master Problem}
The RMP is initialized with artificial slack variables covering each activity with a high penalty cost because initial feasible primal and dual solution are necessary to initiate column generation. The RMP formulation is the same as the one presented in the section above. The only difference is the relaxation of the binary variables. The RMP is solved using Gurobi LP solver. The dual solution is known to be highly unstable over the initial iterations of the column generation leading to poor columns being generated. To tackle such instability, different stabilization methods like BoxStep method \cite{marsten1975boxstep} and Du Merle \cite{du1999stabilized} method have been developed in the literature. We solve the RMP  with barrier linear programming (LP) method without crossover. Its has been seen across literature that barrier method seems to be more efficient to stabilize the duals. The dual solution is centered since it points to an interior solution.

\subsection{Pricing problems}
We have one pricing sub-problem for each tail.The structure of the pricing problem for each tail is as shown in the Fig. 5. The network for the  pricing problem for a tail is represented by a connection flight network. The source node represents the carry-in flight for the tail which specifies the present availability of the tail at a particular base at a particular period of time. A sink node is added to the network and several restrictions can be imposed through it. If a particular tail has to return to a particular airport in the night, it can be enforced through the sink node. The valid flights that can be qualified to be served by the particular tail and the pre-assigned activities are represented by the nodes in the figure. The flight nodes have inter-arcs between them which represent the flight connections that are possible which are known a priori. In the Fig. 5, the source is connected to flights for which the departure base is same as the arrival base of the carry-in flight. All the flights are connected to the sink. The connections are determined a priori based on several business rules mainly MGT and max connection time. Some business heuristics are helpful in reducing the density of the network without affecting the solution quality. The dual information $\pi$ = \{$\pi_1$,$\pi_2$,...,$\pi_f$\}, $\beta$ = \{$\beta_1$,$\beta_2$,...,$\beta_t$\} where $\pi$ corresponds to the duals related to the activity constraints and $\beta$ corresponds to the tail constraints. They are obtained after solving the RMP and used to solve the pricing problem. \par

\begin{figure}[h]
\centering
\noindent
\includegraphics[width=\textwidth]{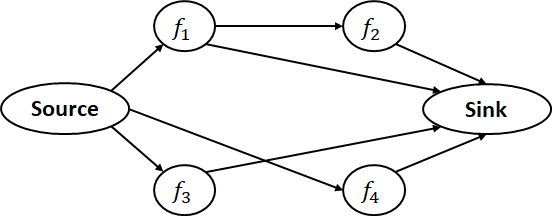}
\caption{flight connection network}
\end{figure}

\noindent
\textbf{Labeling Algorithm} To solve the problem, we employ the label-setting algorithm proposed by Desrochers and Soumis \cite{desrochers1988generalized}. Each label at particular node represent the different paths of reaching the node from the source. A label at a particular node label L =  \{$\Bar{c_j}$,$r^1$,$r^2$,...,$r^k$\} keeps tracks of the information like reduced costs and resource consumption accumulated along its path. In the label, $\bar{c_j}$ is reduced cost of the route and $r_k$ is the consumption of resource k. The labeling algorithm works by pushing labels from the predecessor to all its successor and updating the reduced costs and resources consumptions. The nodes to be treated are ordered according to the starting times of activities and for each node, we push the labels to all its successors. At every node, we retain only few significant labels based on dominance among labels. A label p dominates label q if $\bar{c_p} \leq \bar{c_q}$ , and $r_p^k\leq r_q^k$ for every resource k. So we retain only non-dominated labels at a particular node. In reality, due to large number of resources  we would end up with up large number of non-dominated labels on a particular node. In such cases, we adopt the lexicographical sorting strategy proposed by Gronviskt \cite{gronkvist2005tail} to limit the number of labels stored. Lexicographical ordering limits the number of labels stored at each node when dominance alone is not enough. When the number of labels at a particular node are greater than 12 and less than 20, we employ the lexicographical sorting technique. We do no retain more than 20 labels at each node except the sink node. After all the nodes have been treated, the negative reduced cost paths from the sink node are added to the RMP as variables. These feasible routes are transformed as variables covering certain activities with associated costs and  are added to the  RMP model. \par 

\subsection{Approach}

\noindent
\textbf{Handling Restrictions}. Restrictions include pre-assigned activities, FH\& FC restrictions, and planning maintenance opportunities along the route. First, the pre-assignment activities, which are usually the long term planned maintenance, are ensured by treating them as intermediate sink node and only labels from the intermediate node are extended ahead since they have to be necessarily assigned. The presence of pre-assigned activities further decomposes the shortest path problem. Secondly, the FH and FC restrictions that are  specific to each tail are determined by the M\&E team well ahead of time. These restrictions are handled in the pricing problem by using resource attributes in the label. We keep track of the resources of type k consumed throughout. For each specific tail, the upper bounds on the resource consumption of resource k is different and only legal labels will be extended to the successor that satisfies the upper bounds on the resource consumption. Third, since our problem also includes identifying maintenance opportunities when the aircraft is present at the maintenance hub and sufficient time is available for maintenance before the next flight. However, we can only perform the maintenance activity if there is sufficient capacity available at the hanger. For these reason, we extend two labels from the node to its successor. Considering that the maintenance activity is performed, the first label is extended and the relevant resources are to zero. The second label is extended without maintenance and the accumulation of the resource is retained as shown in Fig.6. We create a maintenance opportunity 'M' between the two flights F2 and F3 to carryout the necessary maintenance activities. The connections time of these two flights should be greater than the required maintenance time and connection base must correspond to one of the maintenance bases.\par

\begin{figure}[h]
\noindent
\includegraphics[width=\textwidth]{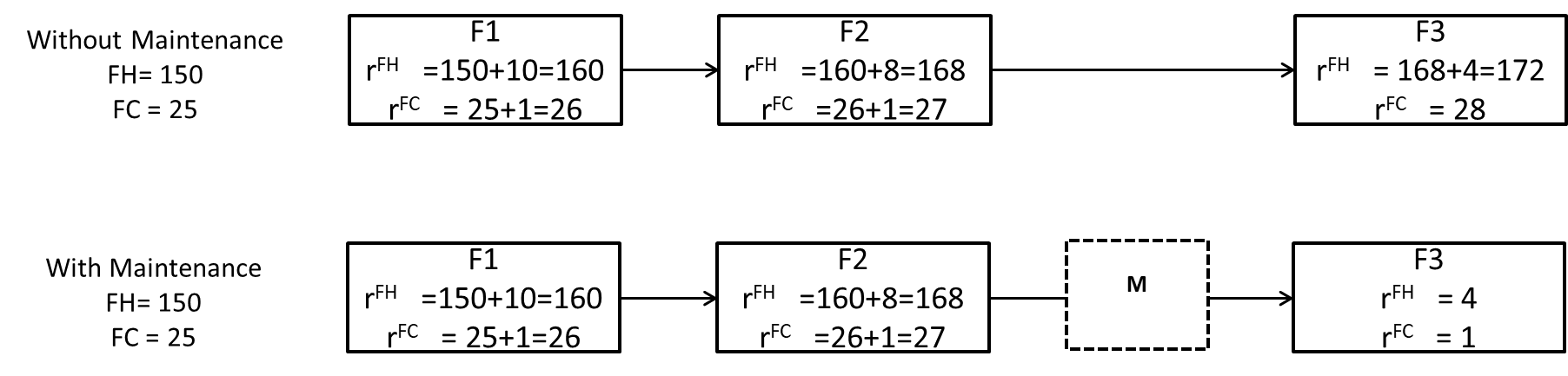}
\caption{Maintenance opportunity example}
\end{figure}

\noindent 
\textbf{Parallel resolution of pricing  problems}. Since all the pricing problems utilize the same dual information, all the pricing problems tend to generate columns covering the same activity which are good from a dual perspective. Eventually the activity will be covered by one single tail in the optimal solution. To overcome this similarity of columns being generated, Gronviskt \cite{gronkvist2005tail} suggests a dual re-evaluation scheme to generate disjoint columns by penalizing the dual values of the activities covered by the previous pricing problems. But this requires the pricing problems to be solved in a serialized order. Re-ordering the pricing problems in a random fashion or ranking them can assist being un-biased to each pricing problem. But instead of serially resolving the pricing problems, we resort to solve the pricing problems in parallel. Since all the pricing problems can be solved independently, time is reduced significantly. Now to overcome the problem of similarity of columns being generated, we retain about 100 negative reduced costs paths for each tail at the sink. From the pool of paths generated from all the pricing problems, we select only few paths which are disjoint in nature and add them to RMP as variables. Based on our study we found that it is good to retain large number of non-dominated labels at sink node for each tail, and later select disjoint paths from the pool. We understand that it is possible that many poor columns are also generated but we add only paths that have a high probability to be part of the solution of RMP. This helps us in solving pricing problems in parallel and keeping the size of RMP less dense. The proposed heuristic for the disjoint path selection is as follows: \par

\begin{algorithm}[H]
\SetAlgoLined
Initialize a  dual vector to retain penalizations: $\pi^{pen}=\{0,...,0\} \hspace{1 mm} , \hspace{1 mm} \emph{C} = \{0,...,0\}$ \\
\For{\texttt{$p \in {\cal{P}}$} paths from solution pool S}{
Let ${\cal{F}}$ be the set of activities covered by path $p$ \\
\eIf{ $\sum_{f \in  {\cal{F}}} \pi_f^{actual} \geq \sum_{f \in  {\cal{F}}} \pi_f^{pen}$ }{
   $C \gets C \cup \{p\}$ ; \\
   \For{\texttt{$f \in {\cal{F}}$}}{
      $\pi_f^{pen}+=\epsilon . \pi_f^{actual}$ ; 
   }
}{
 continue ;
}}
\caption{Disjoint Path selection Heuristic}
\end{algorithm}

\noindent
The value $\epsilon$ can vary between 0.8 to 1. The value 1 means that the selected paths are completely disjoint in nature. Decreasing the value allows the overlapping of activities, thereby more columns are selected from the heuristic. \par

\noindent
\textbf{Integrality}
Even though the column generation technique converges to a better LP solution but still large integrality gap exists. Since routes which are better for an integral solution are yet to be generated. The branch-and-price framework is usually employed to provide better integral solutions for large scale problems. The resolution time of the problem using branch-and-price would be high and also it is very complex in its development. To resolve TAP in a reasonable time, we develop a variable fixing heuristic as shown in Algorithm 2. After the column generation process converges, we initiate the variable fixing heuristic. We always tend to fix variables whose LP solution values are above a certain threshold value. If we find any variables, we fix them to 1 and mark the pricing problems for these tails as inactive. The threshold value changes dynamically throughout the heuristic. When we do not find any variables to fix, we reduce the threshold by 0.05 until we find some variables to fix. Otherwise, even after decreasing the threshold value to an acceptable limit, we exit the fixing loop as no variables were found to fix. Then, we solve the RMP as an integer program with integrality constraints restored. We consider the set $V$ representing the set of variables in RMP  as well as an empty set $F$. We also design three functions. First, \emph{Solve-RMP}, which solves the RMP model and returns the dual solution verctor $\pi$. Second, \emph{Solve-PP($\pi$)}, which resolves all active pricing problems for each tail in parallel and returns a set of generated columns. Third, \emph{DisjointPathSelection($\pi$,$P$)}, which returns a subset of columns from the newly generated columns from PP as per the disjoint path selection heuristic. We fix the initial threshold value to 0.95. 

\begin{algorithm}[H]
\SetAlgoLined
Initialize threshold = 0.95 \\
\While{threshold $\geq$ 0.8}{
 $G=\{\}$ \\
 \For{$v \in {\cal{V\setminus F}}$}{
  \eIf{$v \geq threshold$}{
         $G \gets G \hspace{1mm} \cup \hspace{1mm} \{v\}$
 }{
  continue ;
 }
 }
 
  \eIf{ $G = \emptyset$}{
  $threshold \gets threshold -  0.05$
  }
  {
  threshold = 0.95 \\
  $F \gets F \hspace{1mm} \cup \hspace{1mm} \{G\}$ \\
  Fix the variables in G in the RMP; \\
  $\pi$=Solve-RMP()\\
  $P$=Solve-PP($\pi$)\\
  $S$= DisjointPathSelection($\pi$,$P$)\\
  Add $S$ to RMP, $V \cup \{S\}$ \\ 
  Sovle RMP
  }
}
\caption{Variable fixing heuristic}
\end{algorithm}

\noindent
\textbf{Preprocessing using constraint programming}
It was known from the paper \cite{gronkvist2006accelerating} that simple pre-processing techniques based on Aircraft count balancing do not work in case of flight schedules when there is possibility of unassigned flight activities. Even to make the propagation filter work for CP model, we need to need add the successor of every flight to be itself to allow unassigned activities. This would be very weak in propagation, to handle this \emph{costGCC} propagation is suggested. But the  \emph{costGCC} propagation depends completely on prior knowledge about the upper bound of the solution or number of flights that might be left unassigned. Therefore it would be really difficult to apply it in practice. Therefore we suggest to use the preprocessing only at the initial iterations of the column generation. This would help to accelerate the column generation in the initial stages and later use all the connections for improvement stages. The commercial CP solvers are equipped with only few traditional propagation capabilities like \emph{all-different},  \emph{inverse}, etc. In order to implement certain tunneling constraints for resource consumptions along a path and \emph{costGCC} propagation technique, a specific CP solver must be implemented. Rather than implementing these propagation techniques, we make use of the readily available capabilities of commercial solvers to propagate.

\section{COMPUTATIONAL STUDY}
The test instances  used for the computational studies come from two major air carriers in India. The information related to the instances is  as shown in the Table 6.1. The instances span from 2 to 15 days in window period. The instances are solved at once without considering rolling time slices. The pre-connections refer to all the feasible flight connections. The Post-connections refer to the  connections propagated through the CP model. \par

\begin{table}[h]
\begingroup
\setlength{\tabcolsep}{10pt} 
\renewcommand{\arraystretch}{1.5} 
\centering
\begin{tabular}{c c c c c c}
\toprule
Instance&Horizon(days)&Tails&Flights& Pre-Connections&Post-Connections\\
\midrule
A-120&2&42&509&10057&1969\\
A-129&2&42&539&12000&1333\\
A-45&2&41&563&16404&3196\\
A-3&3&52&955&45401&12903\\
C-5&5&14&243&4270&2175\\
C-7&7&14&335&7982&3349\\
C-10&10&14&470&15369&5201\\
C-15&15&14&715&35324&9364\\
\bottomrule
\end{tabular}%
\endgroup
\caption{Instances information}
\label{table:1}
\end{table}

\noindent For our studies, we consider post-connections only for initial 10 column generation iterations in the network of pricing problems to gain initial speedup. After 10 iterations, we consider all the pre-connections in the the network. The proposed disjoint path selection heuristic in Algorithm 1 significantly reduced the convergence of column generation by about 40\% for the test instances compared to serial resolution of pricing problems. Even though the serial resolution took considerably few iterations to converge than parallel resolution but resolution times were higher. For the serial resolution, we tested the instances based on the ranking order rule as suggested in \cite{gronkvist2005tail}. The comparison between parallel resolution and serial resolution is as shown in Table 6.2.

\begin{table}[H]
\centering
\begingroup
\setlength{\tabcolsep}{10pt} 
\renewcommand{\arraystretch}{1.5} 
\begin{tabular}{c c c c c c}
\toprule
&&\multicolumn{2}{c}{Serial Resolution}&\multicolumn{2}{c}{Parallel Resolution}\\
Instance& LP Obj&Iter&Time(s)&Iter&Time(s)\\
\midrule
A-120&3778764&18&129.7&25&92\\
A-129&3729446&15&96&17&60\\
A-45&1246737&30&990&35&660\\
A-3&2101400&127&6502&133&4640\\
C-5&604633&39&138.7&41&95\\
C-7&704486&144&1014&149&780\\
C-10&727383&168&5049&172&3270\\
C-15&726777&556&38935&582&29950\\
\bottomrule
\end{tabular}%
\endgroup
\caption{Comparison between parallel resolution and serial resolution}
\label{table:1}
\end{table}

\noindent The integral solution to all the test instances with our solution approach is as shown in the Table 6.3. We conducted two set of experiments with and without CP propagation. The rest of the solution approach for both experiment remained same. The solution approach begins with resolving the pricing problems for each tail and converging relaxed LP version of TAP model as in section 4. Then, we initiate fixing the variables by the heuristic presented in Algorithm 2. Later, when no candidate variables are found to fix, we initiate connection fixing. The RMP, an LP program, is later converted to an Integer program by restoring the integrality constraints. 

\begin{table}[H]
\centering
\begingroup
\setlength{\tabcolsep}{10pt} 
\renewcommand{\arraystretch}{1.5} 
\begin{tabular}{c c c c c c }
\toprule
&\multicolumn{2}{c}{Without CP propagation }&\multicolumn{2}{c}{With CP propagation}\\
Instance&Objective&Time(s)&Objective&Time(s)&Remarks\\
\midrule
A-120&3778764&1020&3778764&98.4& Two uncovered flights\\
A-129&3729546&840&3729313&22.8& Two uncovered flights\\
A-45&1252593&3480&1246937&709.2&Complete Assignment\\
A-3&2115250&4707&2115250&2105&One uncovered flights\\
C-5&637041&105&637041&55&Complete Assignment\\
C-7&704486&802&704486&104&Complete Assignment\\
C-10&727355&3760&727355&470&Complete Assignment\\
C-15&74627&29979&745627&3747.37&Complete Assignment\\
\bottomrule
\end{tabular}%
\endgroup
\caption{Instance Results}
\label{table:1}
\end{table}

\noindent The results showed that, using CP propagation at the initial iterations of the column generation, we were able to converge quickly. Since the network problems with CP propagation contain fewer connections, it helps in accelerating the column generation to a better solution. The CP propagation model is resolved considering that there would be no unassigned flights in the final solution which is far from reality, as there is a possibility of flights being left unassigned. So for these reasons, we consider the Post-connections only for 10 initial iterations in our approach.Its evident that with CP propogation we were able to converge to a better solution in less time for all the instances.

\newpage
\section{CONCLUDING REMARKS}
This paper presents a combination of column generation and constraint programming approach supported with efficient heuristics used to tackle the TAP. This problem is mainly related to the operational perturbations that happen in the short term within airline industry. In our paper, we innovatively contributed to the existing literature through an innovative way to ensure the handling of restrictions. We also performed a parallel resolution of the the pricing problems, which proved to be more efficient than the sequential resolution. To do so, we designed an heuristic that deals with the integrality aspect. Finally, we highlighted the role that constraint programming brings to the formulation as well as the impact on the execution time. The tests results proved the effectiveness of our approach in tackling rapid perturbations that happen on the operational level. The paper also provides an exhaustive description of the TAP as well as an updated review of literature related to the optimization in airline problems. The proposed approach has been deployed in production systems at LAI in the Tail Assignment Optimizer (TAO).

\section*{Acknowledgements}
The research work was conducted by the R\&D team of Laminaar Aviation InfoTech. The company provides aviation-specific IT solutions to airlines across the globe. We would also like to extend our gratitude to the LAI team, Mr. Sujayendra Vaddagiri, Mr. Sunil Sindhu and Mr. Mayur Pustode for their continued co-operation and valuable support.

\newpage

\bibliographystyle{unsrt}
\bibliography{references.bbl}
\end{document}